%% file: main.tex
\documentclass[letterpaper, 10 pt, conference]{styles/ieeeconf}
\IEEEoverridecommandlockouts                             
\input{tex/preamble}
\newif\ifanon

\ifanon
\else
\fi

\title{\LARGE \bf HJCD-IK: GPU-Accelerated Inverse Kinematics\\through Batched Hybrid Jacobian Coordinate Descent}

\ifanon
\else
\author{Cael Yasutake$^{1}$, Andrew H. Liu$^{2}$, Zachary Kingston$^{2}$, and Brian Plancher$^{3,4}$%
\thanks{This material is based upon work supported by the National Science Foundation (under Award 2411369). Any opinions, findings, conclusions, or recommendations expressed in this material are those of the authors and do not necessarily reflect those of the funding organizations.}
\thanks{$^{1}$Columbia University, NY, USA. \quad $^{2}$Purdue University, IN, USA.}%
\thanks{$^{3,4}$Barnard College, Columbia University, NY, USA, \& Dartmouth College, NH, USA. Correspondence to:~{\tt\footnotesize plancher@dartmouth.edu}}%
}
\fi

\begin{document}
\maketitle
\thispagestyle{empty}
\pagestyle{empty}

\begin{abstract}
    Inverse Kinematics (IK) is a core problem in robotics, in which joint configurations are found to achieve a (collision-free) desired end-effector pose. Modern IK solvers face a fundamental trade-off: analytical methods are fast but lack generality, while numerical optimization-based methods are broadly applicable but prone to local minima and high computational costs.
    To overcome this challenge, we introduce HJCD-IK, a GPU-accelerated, sampling-based hybrid solver. By pairing a novel orientation-aware greedy coordinate descent initialization with Jacobian-based polishing and a parallel collision filter, our method achieves up to order-of-magnitude gains in speed and accuracy over state-of-the-art solvers, consistently finding collision-free solutions on the accuracy-latency Pareto frontier, while producing a diverse distribution of high-quality samples. We validate our solver on a physical Franka manipulator and release our code open-source.
\end{abstract}

\section{Introduction} \label{sec:intro}
Inverse Kinematics (IK) algorithms compute (collision-free) joint configurations that achieve a desired end-effector pose, a fundamental requirement for numerous robotics applications~\cite{kucuk2004inverse,isaacs1987controlling,aristidou2018inverse,tevatia2000inverse,angeles1985numerical,cohn2024constrained}. While analytical solvers (e.g., IKFast~\cite{diankov2010automated}, IKBT~\cite{zhang2017ikbt}) are computationally efficient, the inherent redundancy of most robots means general closed-form solutions rarely exist, restricting these solvers to low-degree-of-freedom (DoF) systems with specific topologies.

For general or high-DoF manipulators, numerical approaches provide necessary flexibility but incur higher computational costs. Iterative methods popular in computer graphics, such as FABRIK~\cite{aristidou2011fabrik} and Cyclic Coordinate Descent (CCD)~\cite{wang1991combined,welman1993inverse, lander1998making, kulpa2005fast, canutescu2003cyclic, kenwright2012inverse, chen2019general, ojha2023singularity}, yield high-quality solutions but often lack the strict orientation constraint support required for robotics. Consequently, robotics typically relies on first-~\cite{balestrino1984robust, wolovich1984computational, nakamura1986inverse, wampler1986manipulator, buss2005selectively, beeson2015trac} and second-order~\cite{xie2022speedup, goldenberg1985complete} Jacobian-based optimization, which can readily incorporate additional constraints~\cite{beeson2015trac,pink,rakita2018relaxedik,dantam2021robust,wang2023rangedik}. While generally effective, these methods are all \emph{local methods}, and do not guarantee that a solution will be found given an arbitrary initialization~\cite{angeles1985numerical}.

To overcome local minima, ``global'' IK solvers~\cite{huang2012particle, rokbani2013inverse,farzan2014parallel,dai2019global,starke2020bio} have been proposed. However, they remain computationally prohibitive for real-time applications or restricted to specific robot topologies. Similarly, learning-based neural network approaches~\cite{demers1991learning,hasan2006adaptive,ren2020learning,bensadoun2022neural,ames2022ikflow} offer fast, parallelizable inference but struggle with precision, often yielding unacceptable errors on the order of centimeters.

Given these challenges and opportunities, \emph{parallel computing} provides promise through the parallel exploration of multiple candidate configurations that can escape local minima, particularly in cluttered or obstacle-rich environments~\cite{126022,594541,gosselin1993parallel,farzan2014parallel,sun2024optimization}. In particular GPU-accelerated IK approaches specifically aimed at robotics~\cite{6696709,126022,ames2022ikflow,sundaralingam2023curobo,danaci2023inverse,pyroki2025} have been used to produce state-of-the-art results.

Inspired by these recent advances, in this work, we introduce Hybrid Jacobian Coordinate Descent IK (HJCD-IK), a GPU-accelerated hybrid three-phase sampling-based IK solver that supports both self and environment collision avoidance. HJCD-IK combines a novel orientation-aware, greedy coordinate descent initialization scheme, that provides fast, diverse seeds, with a parallel Jacobian-based polishing routine, and a GPU-accelerated collision detection filter. 

\input{tex/figs/figs-fig1}

As previewed in Fig.~\ref{fig:f1}, and detailed in Section~\ref{sec:results}, this design improves both convergence speed and accuracy over the state-of-the-art methods, both learning- and non-learning-based, consistently finding solutions along the accuracy-latency Pareto frontier and often achieving order-of-magnitude gains. In addition, our method produces not only one, but a broad distribution of high-quality samples, yielding the lowest maximum mean discrepancy compared to baseline approaches.
We deploy our solver onto a physical Franka manipulator for real-world demonstrations of real-time, collision-free inverse kinematics (\cref{fig:hardware},~\cref{sec:results:hardware}) and release our solver open-source at:
\ifanon
\url{https://anonymous.4open.science/r/HJCD-IK-43DD}%
\else
\url{https://github.com/a2r-lab/HJCD-IK}.
\fi%

\section{Background} \label{sec:background}

Forward kinematics (FK) determines the pose of a robot manipulator's end-effector given a joint configuration. For an $n$-DoF manipulator, the FK function can be expressed as follows, where $\boldsymbol{\theta} = [\theta_1, \theta_2, ..., \theta_n]$ is the joint configuration, and $\textbf{P}_{ee} \in \text{SE(3)}$ is the end-effector workspace pose:
\begin{equation} \label{eq:fk}
   \textbf{P}_{ee} = f(\theta_1, \theta_2, ..., \theta_n).
\end{equation}
Inverse kinematics (IK) seeks to determine the inverse, the joint angles $\boldsymbol{\theta}^*$ that result in a target pose $\textbf{P}_t \in \text{SE(3)}$:
\begin{equation} \label{eq:ik}
   \boldsymbol{\theta}^* = f^{-1}(\textbf{P}_t)
\end{equation}
Numerical IK solvers generally turn the IK problem into a constrained optimization problem, minimizing the distance between the end-effector pose $\textbf{P}_{ee}$ and the target pose $\textbf{P}_t$, subject to joint limit constraints defined by $\boldsymbol{\theta}_{min}$, $\boldsymbol{\theta}_{max}$, and other constraints (e.g., obstacles, task goals), $g(\boldsymbol{\theta})$:
\begin{equation} \label{eq:opt}
\begin{split}
   \min_{\boldsymbol{\theta}} \quad &||\textbf{P}_{ee}(\boldsymbol{\theta}) - \textbf{P}_t||^2 \\
   s.t. & \quad \boldsymbol{\theta}_{min} \leq \boldsymbol{\theta} \leq \boldsymbol{\theta}_{max}, \quad g(\boldsymbol{\theta}) \geq 0.
\end{split}
\end{equation}

Popular methods for solving~\cref{eq:opt}, e.g., Jacobian-based methods~\cite{10.5555/533662, balestrino1984robust, wolovich1984computational, nakamura1986inverse, wampler1986manipulator, buss2005selectively, beeson2015trac}, operate by iteratively linearizing and minimizing the task residual $r(\boldsymbol{\theta})$, which is the position and orientation error from the target end-effector pose: 
\begin{equation} \label{eq:task_residual}
    r(\boldsymbol{\theta}) = \begin{bmatrix} r_p(\boldsymbol{\theta}) \\ r_o(\boldsymbol{\theta})\end{bmatrix} = \begin{bmatrix}
    \textbf{P}_t - \textbf{P}_{ee}(\boldsymbol{\theta}) \\ \omega(\boldsymbol{\theta}) \end{bmatrix},
\end{equation}
where $\omega(\boldsymbol{\theta})$ encodes the quaternion difference:
\begin{equation} \label{eq:quat_diff}
    \begin{aligned}
        q_{err} &= q_t \otimes q_e^{-1} = [w, \mathbf{v}], \\
        \omega(\boldsymbol{\theta}) &= \frac{2 \arctan2(\|\mathbf{v}\|, |w|)}{\|\mathbf{v}\|}\mathbf{v}.
    \end{aligned}
\end{equation}

This results in a least-squares update to $\boldsymbol{\theta}$~\cite{deo1995overview}. To improve the robustness of this approach near singularities and joint limits, this system is usually solved with a Levenberg--Marquardt (LM) formulation~\cite{sugihara2011solvability}:
\begin{equation} \label{eq:levenberg-marquardt}
    [J(\boldsymbol{\theta})^\intercal J(\boldsymbol{\theta}) + \lambda D]\Delta \boldsymbol{\theta} = J(\boldsymbol{\theta})^\intercal r(\boldsymbol{\theta}),
\end{equation}
where $\lambda > 0$ is a damping factor, $D$ is a positive diagonal matrix that scales joint updates to stabilize the step, and $J(\boldsymbol{\theta})$ is the manipulator Jacobian at the current joint configuration. We note that the columns of $J(\boldsymbol{\theta})$ are as follows (for joint $i$ with world-frame axis $z_i(\boldsymbol{\theta})$ and position $P_i(\boldsymbol{\theta})$):
\begin{equation} \label{eq:jacobian}
    J(\boldsymbol{\theta}) = \begin{bmatrix} z_i(\boldsymbol{\theta}) \times (\textbf{P}_{ee}(\boldsymbol{\theta}) - \textbf{P}_i(\boldsymbol{\theta}))\\ z_i(\boldsymbol{\theta}) \end{bmatrix}_{i=1...N}.
\end{equation}

Finally, since multiple solutions often exist for a given target pose, it would be ideal to find the \emph{global} optimal solution. While this is often computationally infeasible in practice, by leveraging \emph{multiple local} optima, we can enhance overall IK solver performance for real-time robotic systems. This strategy has led to the current state-of-the-art GPU-accelerated solvers~\cite{sundaralingam2023curobo,pyroki2025}. For ease of notation later, we refer to such a batch of $M$ solutions $[\boldsymbol{\theta}_1 \ldots \boldsymbol{\theta}_M]$ as $[\Theta]_M$.

\input{tex/figs/figs-ccdAlg}
\input{tex/figs/fig-design}
\subsection{Cyclic Coordinate Descent (CCD)} 
\label{sec:background:ccd_background}
The CCD algorithm (\cref{alg:ccd}), is an iterative heuristic search technique designed for solving the position-only IK problem~\cite{wang1991combined}, that is, solving for the joint angles $\boldsymbol{\theta}$ that produce the desired $x,y,z$ position of the end effector.

In CCD, joints are numbered $i = 1 \text{ to } n$ starting at the base. As shown in \cref{fig:ccd}, at each step of the algorithm a joint, $j$ is selected from $j = n \text{ to } 1$ (from the tip to the root of the kinematic tree). Using joint $j$'s rotational axis, $\vec{r}_j(\boldsymbol{\theta})$, referred to as $\vec{r}_j$ in the following, the desired rotation angle update, $\Delta \theta_j$, is computed by projecting the vectors between the position of the current rotational axis and both the target and end-effector positions onto rotational axis plane:
\begin{equation} \label{eq:vec_proj}
    \begin{aligned}
        &\vec{u}_j = \frac{\textbf{P}_{ee}(\boldsymbol{\theta}) - \textbf{P}_{j}(\boldsymbol{\theta})}{||\textbf{P}_{ee}(\boldsymbol{\theta}) - \textbf{P}_{j}(\boldsymbol{\theta})||}
        &\vec{v}_j = \frac{\textbf{P}_t  - \textbf{P}_{j}(\boldsymbol{\theta})}{||\textbf{P}_t - \textbf{P}_{j}(\boldsymbol{\theta})||} \\
        &\vec{u}^r_j = \left(\frac{\vec{u}_j \cdot \vec{r}_j}{|\vec{r}_j|^2}\right)\vec{r}_j \quad
        &\vec{v}^r_j = \left(\frac{\vec{v}_j \cdot \vec{r}_j}{|\vec{r}_j|^2}\right)\vec{r}_j \\
        &\vec{u}^\text{~proj}_j = \vec{u}_j - \vec{u}^r_j \quad
        &\vec{v}^\text{~proj}_j = \vec{v}_j - \vec{v}^r_j. \\
    \end{aligned}
\end{equation}
These projected vectors can then be used to compute $\Delta \theta_j$:
\begin{equation} \label{eq:theta_proj}
   \Delta \theta_j = \cos^{-1}\left(\frac{\vec{v}^\text{~proj}_j}{||\vec{v}^\text{~proj}_j||} \cdot \frac{\vec{u}^\text{~proj}_j}{||\vec{u}^\text{~proj}_j||}\right)
\end{equation}
Once all $n$ joints have been updated, the algorithm measures the distance between the end-effector and target position, often through the $L_2$ norm, $||\textbf{P}_{ee}(\boldsymbol{\theta}) - \textbf{P}_t||_2^2$, and exits upon $\epsilon$-convergence. However, if the target is out of reach or CCD becomes locked in a singularity~\cite{ojha2023singularity}, the algorithm will continue until it reaches a preset iteration limit. 

\input{tex/figs/figs-finalAlg}
\section{Orientation-Aware CCD} \label{sec:orientation-ccd}
In this section, we develop an orientation-aware CCD scheme leveraging screw theory and manifold optimization to project not only in position space but also in orientation space. To the best of the authors' knowledge, this is the first time that orientation-based projections have been integrated into a CCD framework for robotics (e.g.,~\cite{canutescu2003cyclic} derived custom update steps specifically for protein loop closure).

Given the end-effector quaternion $q_{ee}(\boldsymbol{\theta})$ and the target quaternion $q_t$, the orientation error $q_{err}(\boldsymbol{\theta})$ is defined as the quaternion difference \cref{eq:quat_diff}. We decompose the resulting quaternion error $q_{err} = [w, \boldsymbol{v}]$ into angle-axis form. The rotation angle $\phi$ is obtained from the scalar part of the quaternion error, while the corresponding axis of rotation is obtained by solving for its vector component~\cite{lechuga2022iterative}:
\begin{equation} \label{eq:angle_axis}
    \begin{aligned}
    &\phi = 2 \times \arccos (w)\\
        &[w, \boldsymbol{v}] = \cos\left(\sfrac{\phi}{2}\right) + \hat{a}\sin\left(\sfrac{\phi}{2}\right)
        \;\; \Rightarrow \;\;
        \hat{a} = \frac{\boldsymbol{v}}{\sin\left(\sfrac{\phi}{2}\right)}
    \end{aligned}
\end{equation}
The angle-axis pair $(\phi, \hat{a})$ defines the orientation update. 

For each joint $j$ with axis $\vec{r}_j(\boldsymbol{\theta})$, again referred to as $\vec{r}_j$ for notational simplicity, the update is projected onto the component of $\hat{a}$ aligned with $\vec{r}_j$ and scaled with an annealed factor $\delta(k)$ for each algorithmic iteration $k$:
\begin{equation} \label{eq:ori_update}
    \Delta \theta_j = \delta(k) \operatorname{sgn}(\hat{a}\cdot \vec{r}_j)\phi \quad \text{with } \operatorname{sgn}(0)=0.
\end{equation}
By considering this orientation update with the standard positional CCD step, our method performs greedy joint selection in both position and orientation space, enabling rapid convergence with respect to Euclidean and angular error. As such, this scheme enables a CCD-style algorithm to solve the full IK problem as shown in~\cref{eq:ik}.

\section{The HJCD-IK Algorithm} \label{sec:design}

In this section we present HJCD-IK (\cref{fig:design}, \cref{alg:hjcd}), a GPU-accelerated, sampling-based hybrid solver that couples an orientation-aware greedy coordinate-descent (PO-CCD) initializer with a parallel Jacobian-based polishing stage (PJ-IK), and leverages recent advances in GPU-accelerated collision detection~\cite{huang2026prrtc} to ensure that the returned solution is feasible. We present the full algorithm in~\cref{alg:hjcd} and detail its sub-components in later sections and algorithms. 

At the highest level, the first stage evaluates hundreds to low-thousands of orientation-aware CCD seeds in parallel (\cref{sec:design:po-ccd}) and returns $M$ seeds (\cref{alg:hjcd} Line 1). These are collected and ranked by their residuals, of which the top $K$ candidates are then duplicated with small perturbations to reduce the risk of converging to local minima, producing a batch of $B$ seeds (\cref{alg:hjcd} Lines 3-7). These are further refined by a batched Jacobian solver (\cref{sec:design:pjik}) in stage 2 to return an optimized batch of $B^*$ configurations (\cref{alg:hjcd} Line 8). These are then filtered for collision avoidance (\cref{alg:hjcd} Line 9) to produce the batch $CF$, and then scored, returning the final optimized configuration $\boldsymbol{\theta}^*$ (\cref{alg:hjcd} Line 10) in stage 3. 

We note that alone, orientation-aware CCD cannot converge quickly to high-precision solutions, and batched Jacobian solvers are prone to getting trapped in spurious local minima. However, when combined, orientation-aware CCD generates a diverse set of coarse solutions from many randomized initializations, providing excellent seeds for the Jacobian-based polishing stage, which can then be filtered for collision avoidance. This design enables fast high-quality batch generation and fast post-hoc feasibility checks.
We also note that HJCD-IK is explicitly co-designed for GPUs, with kernels that exploit block-, warp-, and thread-level parallelism to maximize throughput while minimizing synchronization. This algorithm-hardware-software co-design approach enables real-time performance and, as shown in~\cref{sec:results}, produces IK solutions that are both faster and more accurate than state-of-the-art alternatives.

\subsection{Massively-Parallel Orientation-Aware CCD (PO-CCD)} \label{sec:design:po-ccd}
Our co-designed, orientation-aware, greedy CCD stage (\cref{alg:pccd}) efficiently harnesses GPU parallelism by distributing orientation-aware CCD computations across blocks, warps, and threads, enabling the simultaneous processing of hundreds to low-thousands of sampled configuration seeds. At the highest level this is done by launching $M$ parallel blocks that run asynchronously, one for each sample. Within these samples synchronizations are kept to a minimum, and all underlying linear algebra is parallelized across threads. 

As shown in~\cref{alg:pccd} Lines 2-4, the algorithm begins with uniform sampling of $M$ initial joint configuration seeds within the joint limits of each joint.

We then enter the main orientation-aware CCD loop (\cref{alg:pccd} Lines 5-14) where each block uses two warps of threads for each joint. This ensures that each pair of warps is responsible for performing an update for a single joint of a single IK problem, one for orientation and one for position.
At every iteration, these warps calculate coordinate descent updates while holding all other values constant, producing for each joint, in parallel, the orientation and position task residual (\cref{alg:pccd} Lines 6-8).

To determine the best update for each step, these residuals are stored in shared memory, allowing for fast, low-overhead comparisons across each block in which we greedily select both the best orientation and position update and apply it (\cref{alg:pccd} Lines 9-10). If both updates target the same joint, we choose the larger of the two. To reduce the chance of this process jumping to a local minimum, we require a minimum threshold for improvement, $\gamma$, along either the position or orientation task space. If this update fails, we instead randomly perturb the joint configuration. This greedy joint selection process allows us to skip joint updates that will result in no progress (e.g., when the normal axis of the rotation plane is close to that of the vector made between the joint and the target position) and reduce total computation time (\cref{alg:pccd} Lines 11-13). Once a seed satisfies the position and orientation error thresholds (\cref{alg:pccd} Line 14), the parallel loop is broken and all samples are returned.

Because this initializer is very fast at finding reasonably good solutions for large scales of hundreds to thousands of seeds, but can struggle to satisfy extremely tight tolerances, it terminates at a coarse tolerance with only a modest number of iterations, deferring fine accuracy to the downstream Jacobian-based refinement.

\input{tex/figs/figs-finalAlg-modules}
\subsection{Parallel Jacobian IK} \label{sec:design:pjik}
Stage 2 of our algorithm polishes the returned high-quality batch of refined seeds $B$ into an optimized batch for collision filtering $B^*$ through parallel Jacobian IK (PJ-IK) as shown in~\cref{alg:pjik}. Here, each block handles one seed and parallelizes all underlying linear algebra across the warps of threads within that block. As Jacobian calculations are more expensive than CCD calculations, this reduced batch size of tens to low-hundreds of refined seeds $B$ also scales well on modern GPU hardware. To ensure that these calculations are done maximally efficiently, we leverage and extend the GRiD library~\cite{plancher2021accelerating,plancher2022grid} for all Jacobian calculations, which is also designed for overall block-level parallelism and underlying thread-level linear algebra parallelism. In particular, we add support for arbitrary end-effectors to enable real-world robot hardware validation.

For each candidate $b$, we solve~\cref{eq:ik} by first minimizing a weighted task residual, where $W(\boldsymbol{\theta})$ is a diagonal weighting matrix that normalizes the rows of the Jacobian and adaptively scales the translational and rotational components of the residual. Adapting this to the LM formulation,~\cref{eq:levenberg-marquardt}, and dropping the dependence of all variables on $\boldsymbol{\theta}$ for readability, we solve the following, where $D= \text{diag}(J^\intercal J)$:
\begin{equation} \label{eq:normal-eqs}
    (J^\intercal W^\intercal J + \lambda D) \Delta \boldsymbol{\theta} = -J^\intercal W^\intercal W r,
\end{equation}
The interpolation between the gradient and Gauss-Newton step via the $\lambda D$ damping term provides additional robustness near singularities and joint limits. To further stabilize convergence, each update $\Delta\boldsymbol{\theta}$ is constrained by a trust region of radius $R$, preventing aggressive steps far from the solution. Candidate steps are also validated through a backtracking line-search for the scaling factor $\alpha$, accepting the first iterate that reduces the task residual. This LM step is shown in~\cref{alg:pjik} Lines 3-8. We note that in general, such a line search will solve for $\alpha^*$ when given an update, $\Delta \boldsymbol{\theta}$, original value, $\boldsymbol{\theta}$, and line-search range $\mathcal{A} = [1, \sfrac{1}{\beta}, \ldots, \sfrac{1}{\beta^A}]$, under a cost function, $r(\cdot)$:
\begin{equation} \label{eq:line-search}
\begin{split}
    \alpha^* = \min \{ \alpha_i \mid r(\boldsymbol{\theta} + \alpha_i \Delta\boldsymbol{\theta}) < r(\boldsymbol{\theta}), \alpha \in \mathcal{A} \} \text{ or } 0\\
\end{split}
\end{equation}

If no scaled LM step can be found, we resort to two fallback strategies~\cite{martinez2013towards}. The first, a dogleg method~\cite{mizutani1999powell}, shown in~\cref{alg:pjik} lines 9--11, constructs a step $\Delta\boldsymbol{\theta}(\tau)$ that interpolates between the steepest gradient descent step and the Gauss-Newton step by selecting a value $\tau \in [0, 1]$ such that the update lies within the trust-region $R$~\cite{rosen2014rise}:
\begin{equation} \label{eq:dogleg}
\begin{split}
    &\Delta\boldsymbol{\theta}_{GD} = -\alpha J^\intercal r,
    \quad \quad 
    \Delta\boldsymbol{\theta}_{GN} = -(J^\intercal J)^{-1} J^\intercal r,\\
    &\Delta\boldsymbol{\theta}(\tau) = \tau\Delta \boldsymbol{\theta}_{GD} + (1-\tau) \Delta\boldsymbol{\theta}_{GN},\\
    &||\Delta\boldsymbol{\theta}(\tau)||_2^2 \leq R^2.
\end{split}
\end{equation}
If the dogleg also fails, a single-coordinate line search, shown in~\cref{alg:pjik} Lines 12-14, is attempted by updating the single joint with largest gradient magnitude according to the norm of the weighted residual~\cite{wright2015coordinate}:
\begin{equation} \label{eq:single-coord}
    \Delta\theta = \max\limits_i \frac{\partial}{\partial \theta_i} \frac{1}{2}||W(\boldsymbol{\theta})r(\boldsymbol{\theta})||^2.
\end{equation}
If all of these fail, a random perturbation is made to avoid local minima (\cref{alg:pjik} Line 15). This is repeated until convergence or maximum iterations (\cref{alg:pjik} Line 16). These combined refinement stages are able to consistently return precise solutions from the coarse candidates, achieving sub-millimeter positional and sub-degree rotational accuracy.

\subsection{GPU-Parallel Collision Checking} \label{sec:design:cc}
The final algorithm stage is a GPU-parallel collision avoidance filter. Here we leverage similar block-, warp-, and thread-optimized approaches as with our prior two stages, adapted from~\cite{huang2026prrtc}. Offline, we use the \texttt{foam} tool~\cite{coumar2025foam} to generate bounding spheres for the robot geometry. Online, we check for collisions against those spheres through a two-stage hierarchical check: a low-resolution pass flags potential collisions, followed by a high-resolution pass on flagged links for precision. To maximize throughput, groups of four threads execute 4x4 matrix operations for forward kinematics, while warp primitives enable low-latency ``early exits'' upon collision detection. This architecture enables fast queries that scale efficiently across the tens to low hundreds batch sizes common for $B^*$. We refer the reader to~\textcite{huang2026prrtc} for more details on the underlying kernel design.

\section{Results} \label{sec:results}
We benchmark HJCD-IK against three other GPU-parallel IK solvers: cuRobo~\cite{sundaralingam2023curobo}, IKFlow~\cite{ames2022ikflow}, and PyRoki~\cite{pyroki2025}. These solvers represent different state-of-the-art approaches for solving IK problems based on: GPU-accelerated optimization, generative modeling with normalizing flows, and differentiable JAX-based computation, respectively.
All results were collected using a laptop with an Intel Core i7-14700HX CPU (20 core, 2.1 GHz base), an NVIDIA GeForce RTX 4060, WSL Ubuntu 24.04, and CUDA 12.5. We benchmark in simulation the 7-DoF Franka Panda~\cite{haddadin2022franka} and Fetch~\cite{wise2016fetch} arms. We evaluate open-world poses using joint configuration samples from a Halton Sequence to generate 100 feasible target poses. We evaluate collision avoidance using end-effector poses from the \emph{box\_panda} scene in MotionBenchMaker~\cite{chamzas2021motionbenchmaker}. 
Hardware experiments use a 7-DoF Franka Research 3~\cite{FrankaResearch3}.

\subsection{Batch-Size Scalability} \label{sec:results:bs-scalability}

We first present a scalability study computing open-world (\cref{tab:ik_batch_panda_fetch}) and collision-free (\cref{fig:meta_colfree},~\cref{tab:ik_colfree_box}) motions with sample batch sizes of $B\in \{1, 10, 100, 1000, 2000\}$.
Across both robots and scenarios, HJCD-IK benefits from increased batch size, yielding rapid error reductions with only modest per-target latency increases. This results in our approach (shown in orange) outperforming state-of-the-art baselines in terms of latency across all batch sizes, while also generally surpassing all baselines in accuracy, remaining on or near the accuracy-latency Pareto frontier.

In open-world benchmarks, increasing batch size reduces Panda position and orientation errors by three to four orders of magnitude, respectively, with Fetch showing even greater improvements. Crucially, HJCD-IK incurs minimal latency penalties with solve times increasing only slightly (4.04ms to 4.37ms for Panda; 2.59ms to 2.73ms for Fetch), yielding order-of-magnitude speedups over IKFlow and ${>3.1}$x over PyRoki and cuRobo. Collision-free benchmarks on Panda follow this trend, with error reductions of five to six orders of magnitude and latency varying only from 4.19ms to 5.44ms, with order-of-magnitude speedups over PyRoki and cuRobo. Overall, once batch size is 100 or grater, HJCD-IK is the fastest and most accurate solver, and achieves 100\% success rate on the challenging \emph{box\_panda} task.

\input{tex/figs/figs-batchTable}
\input{tex/figs/fig-colfreeTable}
\input{tex/figs/fig-dofTable}
\input{tex/figs/fig-colFreeStudy}
\input{tex/figs/fig-dofScalingStudy}
\subsection{DoF Scalability} \label{sec:results:dof-scalability}
We assess scalability with respect to manipulator complexity by fixing the batch size to $B = 1000$ and evaluating 7-, 12-, 18-, and 24-DoF Panda Arm variants\footnote{We create the 12-, 18-, and 24-DoF variants using chained copies of the first 6 joints and links of the Panda.} using HJCD-IK, cuRobo, and PyRoki in an open-world environment. As shown in \cref{fig:meta_dof} and~\cref{tab:ik_dof}, HJCD-IK maintains the lowest pose error at every DoF, with position confined to $1.71\times 10^{-5}$ to $3.84 \times 10^{-5}$\,mm and orientation error to $4.11\times 10^{-8}$ to $7.32 \times 10^{-8}$\,rad, indicating added redundancy does not inflate error. Notably, as DoF increases, this error improvement grows to multiple orders-of-magnitude over baselines. HJCD-IK also demonstrates immense scalability, incurring minimal latency penalties over increases in DoFs. HJCD-IK is consistently faster while maintaining better accuracy compared to both solvers, presenting speedups ranging from $3.55$x--$5.22$x over PyRoki and $2.14$x--$4.19$x over cuRobo. Overall, HJCD-IK remains on the accuracy-latency Pareto frontier across all DoFs, showcasing faster IK solve times while converging to lower errors.

\subsection{Solution Space Distribution} \label{sec:results:sol-space-dist}
Returning batches of diverse solutions indicates that a batched local solver is doing a good job at sampling across the space of local optima to find good globalizing solutions.
To evaluate this, we computed the Maximum Mean Discrepancy (MMD) between solver joint configurations and an approximate ground-truth distribution generated via TRAC-IK~\cite{beeson2015trac}. For each of 100 target poses, we compared the 50 best configurations from a batch of 2000 solves from each solver against 50 randomly seeded ground-truth samples. As shown in Table~\ref{tab:mmd}, HJCD-IK achieves the lowest MMD (0.02261) and squared MMD (0.00051), indicating that its returned solutions most closely approximate the ground-truth manifold. While IKFlow is competitive in MMD (0.03670), both PyRoki and cuRobo exhibit significantly higher scores, suggesting reduced diversity. The superior performance of HJCD-IK, particularly in $\text{MMD}^2$, demonstrates its ability to consistently preserve solution redundancy and maintain high-variance coverage even with a minimized return batch. We diagram the impact of this in~\cref{fig:f1}, demonstrating the larger breadth of solutions returned by HJCD-IK over baselines.

\subsection{Hardware Deployment} \label{sec:results:hardware}
We demonstrate HJCD-IK's real-time capabilities, solution diversity, and collision avoidance on a 7-DoF Franka Research 3 manipulator. 
With every camera frame, we obtain both the target and obstacle poses and give them to HJCD-IK, using a batch size of 1000 to obtain a diverse set of collision-free configurations on the self-motion manifold. To ensure sufficiently smooth motion, we choose to move towards the solution that is nearest to the current configuration.

Through this demonstration, we show the robustness and real-world applicability of HJCD-IK for accurate tracking in the presence of moving obstacles. Additionally, the diversity of solutions offered by HJCD-IK ensures that we are able to generate continuous, collision-free motions, which will be useful in future applications to pathwise IK problems.

\section{Conclusion and Future Work} \label{sec:conclusion}
HJCD-IK provides fast, accurate, collision-free IK solutions for any kinematically redundant manipulator operating in SE(3) by combining a massively parallel, orientation-aware greedy coordinate descent initialization, with a parallel Jacobian-based polishing scheme, and a parallel collision filter. Our experiments show that HJCD-IK outperforms the state-of-the-art methods, remaining on or near the accuracy-latency Pareto frontier across batch sizes and robot DoFs, resulting in order-of-magnitude improvements in latency or accuracy. HJCD-IK also provides lower MMD and MMD$^2$ scores for batches of solutions, indicating greater solution diversity and coverage. Finally, we deploy our solver for real-world demonstrations of real-time, collision-free inverse kinematics and release our solver
open-source.

Looking ahead, there are several promising directions for future work. One of particular note is that while our empirical results demonstrate HJCD-IK's strong performance, formal analysis of its convergence and optimality remains an important open area of research. 

\input{tex/figs/fig-hardware}
\input{tex/figs/fig-mmdTable}

\printbibliography{}

\end{document}

%% file: tex/preamble.tex
\usepackage[utf8]{inputenc}
\usepackage[T1]{fontenc}
\usepackage[english]{babel}

\usepackage{xurl}
\usepackage{color}
\usepackage{wrapfig}
\newif\iffigs
\figstrue 

\usepackage{tabularx}
\usepackage{multirow}
\usepackage{booktabs}
\usepackage{colortbl}
\usepackage{siunitx}

\usepackage{etoolbox}
\makeatletter
\patchcmd{\@makecaption}
{\scshape}
{}
{}
{}
\makeatother

\usepackage{subcaption}
\usepackage[fleqn]{amsmath}
\usepackage{float}
\usepackage{setspace}
\usepackage{graphicx}
\usepackage{mathrsfs}
\usepackage{amssymb}
\usepackage{nicefrac}
\usepackage{algorithm}
\usepackage[noend]{algpseudocode}

\makeatletter
\newcommand\fs@spaceruled{\def\@fs@cfont{\bfseries}\let\@fs@capt\floatc@ruled
  \def\@fs@pre{\vspace{0.4\baselineskip}\hrule height.8pt depth0pt \kern2pt}%
  \def\@fs@post{\vspace{-0.4\baselineskip}\kern2pt\hrule\relax\vspace{-12pt}}%
  \def\@fs@mid{\kern2pt\hrule\kern2pt}%
  \let\@fs@iftopcapt\iftrue}
\makeatother
\usepackage{tikz}
\usetikzlibrary{decorations.pathreplacing,calc}
\newcommand{\tikzmark}[1]{\tikz[overlay,remember picture] \node (#1) {};}
\newcommand*{\AddNote}[7]{%
    \begin{tikzpicture}[overlay, remember picture]
        \draw [decoration={brace,amplitude=0.5em},decorate,thick,black]
            ($(#3)!(#1.north)!($(#3)-(0,1)$)$) --  
            ($(#3)!(#2.south)!($(#3)-(0,1)$)$)
                node [align=center, text width=#7cm, pos=#6, anchor=#5, text=black] {#4};
    \end{tikzpicture}
}
\newcommand*{\AddNoteLeft}[5]{
    \begin{tikzpicture}[overlay, remember picture]
        \draw [decoration={brace,amplitude=0.5em,mirror},decorate,thick,black]
            ([xshift=-#5] #1.north) -- ([xshift=-#5] #1.north |- #2.south)
                node [midway, left=0.3cm, align=center, text width=#4cm] {#3};
    \end{tikzpicture}
}

\usepackage{listings}
\lstnewenvironment{itemlisting}[1][]
 {%
  \mbox{}
  \vspace*{-\baselineskip}
  \lstset{
    xleftmargin=\leftmargin,
    linewidth=\linewidth,
    #1
  }%
 }
 {}
\usepackage{flushend}
\usepackage{multicol}

\usepackage{lipsum}

\usepackage{lipsum}
\usepackage{csquotes}
\usepackage[
	maxbibnames=99,
	maxcitenames=2,
	style=numeric-comp,
	backend=biber,
	sorting=none,
	giveninits=true,
	url=false,
	doi=false,
	eprint=false,
	isbn=false,
]{biblatex}

\usepackage[pdfa,colorlinks,bookmarksopen,bookmarksnumbered,allcolors=black,urlcolor=blue]{hyperref}

\usepackage[nameinlink,capitalise]{cleveref}
\crefname{line}{line}{lines}
\crefname{figure}{Fig.}{Figs.}
\Crefname{figure}{Fig.}{Figs.}
\crefname{equation}{Eq.}{Eqs.}
\Crefname{equation}{Eq.}{Eqs.}
\crefname{section}{Sec.}{Secs.}
\Crefname{section}{Sec.}{Secs.}
\crefname{definition}{Def.}{Defs.}
\Crefname{definition}{Def.}{Defs.}
\crefname{algorithm}{Alg.}{Algs.}
\Crefname{algorithm}{Alg.}{Algs.}
\crefname{assumption}{Asm.}{Asms.}
\Crefname{assumption}{Asm.}{Asms.}
\crefname{subassumption}{Asm.}{Asms.}
\Crefname{subassumption}{Asm.}{Asms.}
\Crefname{problem}{Problem}{Problems}
\crefname{problem}{Problem}{Problems}

\usepackage{xfrac}

%% file: tex/figs/figs-fig1.tex
\begin{figure}
    \centering
    \includegraphics[width=1\linewidth]{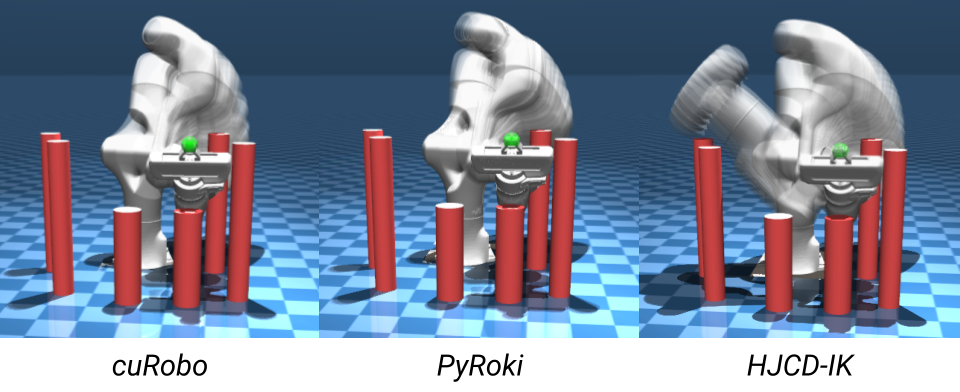}
    \caption{Distribution of collision-free inverse kinematics (IK) solutions. Comparison between cuRobo (left), PyRoki (center), and HJCD-IK (right) for a representative target pose. By leveraging GPU-accelerated parallelism through a three-phase initialization, refinement, and collision filtering regime, HJCD-IK produces a broader distribution of locally-optimal solutions, ultimately yielding a more accurate final result in less time.}
    \label{fig:f1}
    \vspace{-15pt}
\end{figure}

%% file: tex/figs/figs-ccdAlg.tex
\iffigs
\begin{figure}[!t]
    \centering
    \vspace{3pt}
    \includegraphics[width=0.6\columnwidth]{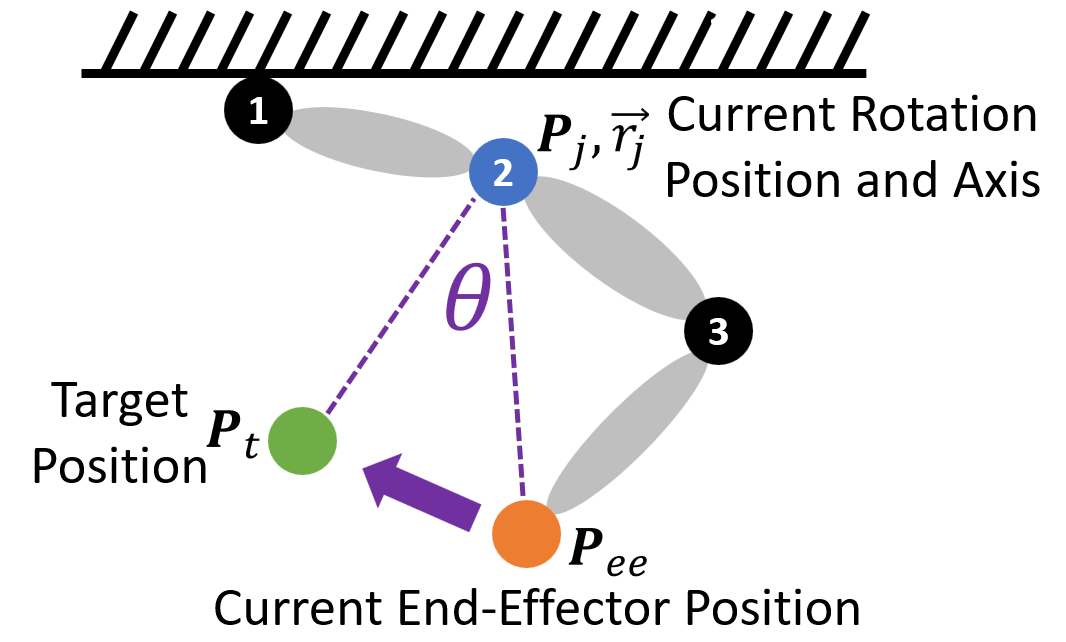}
    \caption{Illustration of the CCD algorithm showing the position of the current end-effector ($P_{ee}$), current rotation joint ($P_j$), and target ($P_t$).}
    \label{fig:ccd}
    \vspace{-10pt}
\end{figure}
\fi
\floatstyle{spaceruled}
\restylefloat{algorithm}
\begin{algorithm}[!t]
\begin{spacing}{1.25}
\begin{algorithmic}[1]
\caption{CCD ($\boldsymbol{\theta}, \epsilon, \text{max\_iters}$ $\rightarrow$ $\boldsymbol{\theta}^*$)}
\label{alg:ccd}
\For{iteration $k$ in max\_iters}
    \For{joint $j$ in $J$ from $n:1$}
        \State Compute $\vec{u}_{proj}, \vec{v}_{proj}$ by \cref{eq:vec_proj} \label{line:compute_vec_proj}

        \State Compute $\Delta \theta_j$ by \cref{eq:theta_proj} \label{line:compute_theta}
        
        \State $\theta_j \gets \theta_j + \Delta \theta_j$ \label{line:joint_update}
    \EndFor
    \State Compute $\textbf{P}_{ee}(\boldsymbol{\theta})$ by \cref{eq:fk}
    \If{$||\textbf{P}_{ee}(\boldsymbol{\theta}) - \textbf{P}_t||_2^2 < \epsilon$} \textbf{break}
    \EndIf
\EndFor
\State \textbf{return} $\boldsymbol{\theta}$
\end{algorithmic}
\end{spacing}
\end{algorithm}

%% file: tex/figs/fig-design.tex
\iffigs
\begin{figure*}[!t]
   \centering
   \vspace{5pt}
   \includegraphics[width=\textwidth]{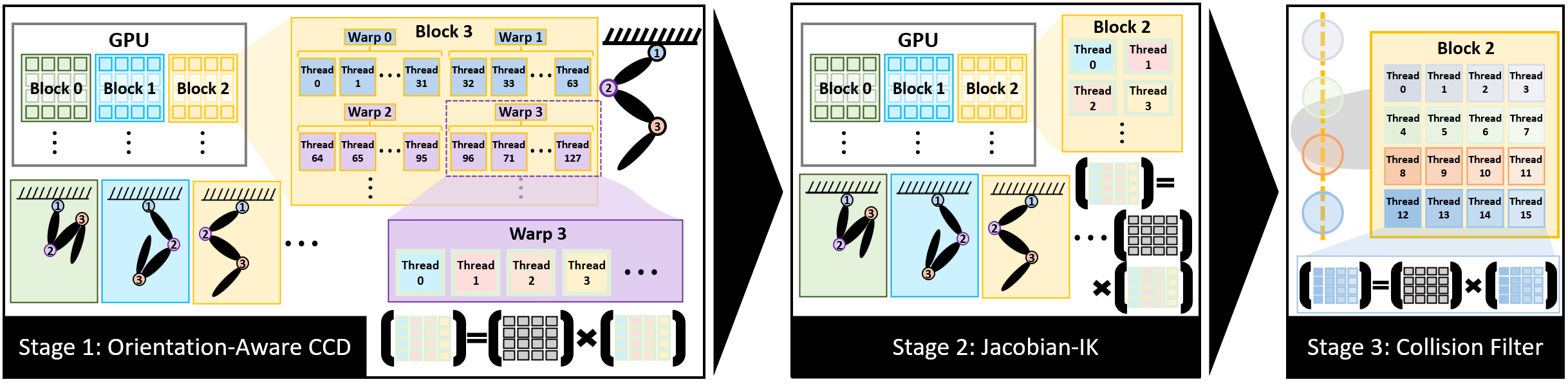}
   \caption{HJCD-IK leverages the large-scale parallelism available on modern GPUs through a three-stage process. First, block-level parallelism is exploited across hundreds to low-thousands of candidate initializations. Each is optimized through warp-level parallelism per joint to compute the position and orientation updates respectively within each CCD phase. The underlying linear algebra for each computation is parallelized across threads. Second, tens to low-hundreds of high-quality candidates are further refined through block-level parallelism via a Jacobian-based polishing routine. Third, we adapt the collision avoidance approach of~\cite{huang2026prrtc} to filter our candidate solutions for feasibility. We note that both stages 2 and 3 also leverage warp- and thread-level parallelism for their underlying linear algebra and Jacobian computations.}
   \label{fig:design}
   \vspace{-15pt}
\end{figure*}
\fi

%% file: tex/figs/figs-finalAlg.tex
\floatstyle{spaceruled}
\restylefloat{algorithm}
\begin{algorithm}[!t]
\begin{spacing}{1.25}
\begin{algorithmic}[1]
\caption{HJCD-IK ($M,K,B,\mu,\Sigma \rightarrow \boldsymbol{\theta}^*$)}
\label{alg:hjcd}
\State Compute $[\Theta]_M = [\boldsymbol{\theta}^0 \ldots \boldsymbol{\theta}^M]$ via \cref{alg:pccd}
\State $[\Theta]_K, [\Theta]_B \gets \emptyset$
\For{$i = 1\ldots K$}
    \State $\boldsymbol{\theta}_i = \text{arg}\min\limits_{[\Theta]_M} r(\boldsymbol{\theta})$ 
    \State $[\Theta]_K \gets [\Theta]_K \cup \boldsymbol{\theta}_i$, \quad $[\Theta]_M \gets [\Theta]_M \setminus \boldsymbol{\theta}_i$
\EndFor
\For{$i = 1\ldots \left \lfloor{B/K}\right \rfloor$}
    \State $[\Theta]_B \gets [\Theta]_B \cup [\Theta]_K + \mathcal{N}(\mu, \Sigma)$
\EndFor
\State Compute $[\Theta]_{B^*}$ via \cref{alg:pjik}
\State Filter $[\Theta]_{B^*}$ for collisions to produce $[\Theta]_{CF}$
\State \Return $\boldsymbol{\theta^*} \gets \text{arg}\min\limits_{[\Theta]_{CF}} r(\boldsymbol{\theta})$
\end{algorithmic}
\end{spacing}
\end{algorithm}

%% file: tex/figs/figs-finalAlg-modules.tex
\floatstyle{spaceruled}
\restylefloat{algorithm}
\begin{algorithm}[!t]
\begin{spacing}{1.25}
\begin{algorithmic}[1]
\caption{PO-CCD ($M, \epsilon, \nu, \gamma, \mu, \Sigma, \text{max\_iters} \rightarrow [\Theta]_M$)}
\label{alg:pccd}
\For{$m = 0 \ldots M$} \textbf{in parallel blocks}
    \For{joint $j$ in $m$} \textbf{in parallel threads} 
        \State $\theta_j^m \sim U(\boldsymbol{\theta}_{min}, \boldsymbol{\theta}_{max})$
        \State Compute initial $r(\theta_j^m)$ by \cref{eq:task_residual}
    \EndFor \tikzmark{bottomr}
    \For{iteration $k$ in max\_iters}
        \For{joint $j$ in $m$} \textbf{in parallel warps}
            \State Compute $\Delta \theta_j^m$ by \cref{eq:vec_proj}, \eqref{eq:theta_proj} or \eqref{eq:angle_axis}, \eqref{eq:ori_update}
            \State Compute $r(\theta_j^m + \Delta \theta_j^m)$ by \cref{eq:task_residual}
        \EndFor
        \State $[\Delta \theta^{m*}_p, \Delta \theta^{m*}_o] \gets \text{arg}\min\limits_{j} r_{[p,o]}(\theta_j^m + \Delta \theta_j^m)$
        \State $\boldsymbol{\hat{\theta}}^m \gets \boldsymbol{\theta}^m + \Delta \theta^{m*}_{[p,o]}$
        \If{$||r_{[p,o]}(\boldsymbol{\hat{\theta}}^m) - r_{[p,o]}(\boldsymbol{\theta}^m)||_2^2 > \gamma$}
            \State $\boldsymbol{\theta}^m \gets \boldsymbol{\hat{\theta}}^m$
        \Else~$\boldsymbol{\theta}^m \gets \boldsymbol{\theta}^m + \mathcal{N}(\mu,\Sigma)$
        \EndIf
        \If{$||r_{[p,o]}(\boldsymbol{\theta}^m)||_2^2 < [\epsilon,\nu]$} \textbf{break}
        \EndIf
    \EndFor
\EndFor
\State \Return $[\Theta]_M = [\boldsymbol{\theta}^0 \ldots \boldsymbol{\theta}^M]$
\end{algorithmic}
\end{spacing}
\end{algorithm}

\floatstyle{spaceruled}
\restylefloat{algorithm}
\begin{algorithm}[!t]
\begin{spacing}{1.25}
\begin{algorithmic}[1]
\caption{PJ-IK ($B, R, \varepsilon, \upsilon, \mu, \Sigma, \text{max\_iters}$ $\rightarrow [\Theta]_{B^*}$)}
\label{alg:pjik}
\For{sample $b = 0 \ldots B$} \textbf{in parallel blocks}
    \For{iteration $k$ in max\_iters}
        \For{joint $j$ in $b$} \textbf{in parallel warps}  \tikzmark{topLM} \tikzmark{sideLM}
            \State Construct $J_j(\theta^b)$ \eqref{eq:jacobian}
        \EndFor
        \State Compute $\Delta\boldsymbol{\theta}^b$ by \eqref{eq:normal-eqs}
        \State $\Delta\boldsymbol{\theta}^b \gets \max(-R, \min(\Delta\boldsymbol{\theta}^b, R))$
        \State Compute $\alpha^*$ for $\Delta\boldsymbol{\theta}^b$ by \eqref{eq:line-search}
        \If{$\alpha^* > 0$} $\boldsymbol{\theta}^b \gets \boldsymbol{\theta}^b + \alpha^*\Delta \boldsymbol{\theta}^b$ \tikzmark{bottomLM}
        \Else\tikzmark{sideDogL}~Compute $\Delta\boldsymbol{\theta}(\tau)$ by \eqref{eq:dogleg} \tikzmark{topDog}
            \If {$r(\boldsymbol{\theta}^b + \Delta\boldsymbol{\theta}(\tau)) < r(\boldsymbol{\theta}^b)$} \tikzmark{sideDog}
                \State $\boldsymbol{\theta}^b \gets \boldsymbol{\theta}^b + \Delta\boldsymbol{\theta}(\tau)$ \tikzmark{bottomDog}
            \Else ~Compute $\Delta \theta$ by \eqref{eq:single-coord} \tikzmark{topSingle}
                \State Compute $\alpha^*$ for $\Delta\theta$ by \eqref{eq:line-search} \tikzmark{sideSingle}
                \If{$\alpha^* > 0$} $\boldsymbol{\theta}^b \gets \boldsymbol{\theta}^b + \alpha^*\Delta\theta$ \tikzmark{bottomSingle}
                \Else~$\boldsymbol{\theta}^b \gets \boldsymbol{\theta}^b + \mathcal{N}(\mu,\Sigma)$ 
                \EndIf
            \EndIf
        \EndIf
        \If{$r(\boldsymbol{\theta}^b) < [\varepsilon,\upsilon]$} \textbf{break}
        \EndIf
    \EndFor
\EndFor
\State \Return $[\Theta]_{B^*} = [\boldsymbol{\theta}^0 \ldots \boldsymbol{\theta}^B]$
\end{algorithmic}
\end{spacing}
\AddNote{topLM}{bottomLM}{sideLM}{LM Step}{west}{0.5}{1}
\AddNote{topDog}{bottomDog}{sideDog}{Dogleg Step}{west}{0.5}{1.2}
\AddNoteLeft{topSingle}{bottomSingle}{Single Coordinate Step}{0.9}{4cm}
\end{algorithm}

%% file: tex/figs/figs-batchTable.tex
\begin{table*}[t]
\centering
\large
\setlength{\tabcolsep}{4pt}
\caption{Batch-size comparison across solvers for the Panda and Fetch Arms in the open-world random target benchmark. Values are the averages across all solves. Position error is in millimeters and orientation error is in radians. HJCD-IK demonstrates as much as order-of-magnitude improvements.}
\label{tab:ik_batch_panda_fetch}
\vspace{-4pt}
\resizebox{\textwidth}{!}{%
\begin{tabular}{r|ccc|ccc|ccc|ccc}
\hline
Batch & \multicolumn{3}{c|}{HJCD-IK} & \multicolumn{3}{c|}{PyRoki} & \multicolumn{3}{c|}{cuRobo} & \multicolumn{3}{c}{IKFlow} \\
\cline{2-13}
& Time (ms) & Pos. Err. (mm) & Ori. Err. (rad)
& Time (ms) & Pos. Err. (mm) & Ori. Err. (rad)
& Time (ms) & Pos. Err. (mm) & Ori. Err. (rad)
& Time (ms) & Pos. Err. (mm) & Ori. Err. (rad) \\
\hline
\multicolumn{13}{c}{\textbf{Panda Benchmarks}} \\
\hline
1    & \cellcolor{blue!10}\textbf{4.04} & $7.04 \times 10^{-2}$ & $2.04 \times 10^{-3}$ & 14.86 & \cellcolor{blue!10}$\mathbf{1.39 \times 10^{-2}}$ & \cellcolor{blue!10}$\mathbf{1.12 \times 10^{-5}}$ & 5.33 & $2.56 \times 10^{1}$ & $1.11 \times 10^{-1}$ & 18.48 & $4.67 \times 10^{0}$ & $2.28 \times 10^{-2}$ \\
10   & \cellcolor{blue!10}\textbf{3.82} & \cellcolor{blue!10}$\mathbf{1.21 \times 10^{-4}}$ & \cellcolor{blue!10}$\mathbf{6.74 \times 10^{-7}}$ & 14.62 & $1.39 \times 10^{-2}$ & $1.12 \times 10^{-5}$ & 5.55 & $2.49 \times 10^{-3}$ & $3.95 \times 10^{-6}$ & 18.95 & $1.38 \times 10^{0}$ & $6.21 \times 10^{-3}$ \\
100  & \cellcolor{blue!10}\textbf{4.07} & \cellcolor{blue!10}$\mathbf{2.25 \times 10^{-5}}$ & \cellcolor{blue!10}$\mathbf{8.95 \times 10^{-8}}$ & 14.20 & $1.39 \times 10^{-2}$ & $1.12 \times 10^{-5}$ & 6.01 & $9.16 \times 10^{-4}$ & $2.83 \times 10^{-6}$ & 22.29 & $5.94 \times 10^{-1}$ & $2.76 \times 10^{-3}$ \\
1000 & \cellcolor{blue!10}\textbf{4.22} & \cellcolor{blue!10}$\mathbf{1.60 \times 10^{-5}}$ & \cellcolor{blue!10}$\mathbf{9.15 \times 10^{-8}}$ & 13.96 & $1.39 \times 10^{-2}$ & $1.12 \times 10^{-5}$ & 19.80 & $3.67 \times 10^{-4}$ & $1.68 \times 10^{-6}$ & 49.78 & $2.06 \times 10^{0}$ & $5.43 \times 10^{-3}$ \\
2000 & \cellcolor{blue!10}\textbf{4.37} & \cellcolor{blue!10}$\mathbf{1.81 \times 10^{-5}}$ & \cellcolor{blue!10}$\mathbf{5.15 \times 10^{-8}}$ & 13.97 & $1.39 \times 10^{-2}$ & $1.12 \times 10^{-5}$ & 30.30 & $2.65 \times 10^{-4}$ & $1.33 \times 10^{-6}$ & 99.98 & $1.92 \times 10^{0}$ & $6.59 \times 10^{-3}$ \\
\hline
\multicolumn{13}{c}{\textbf{Fetch Benchmarks}} \\
\hline
1    & \cellcolor{blue!10}\textbf{2.59} & $5.79 \times 10^{-1}$ & $1.20 \times 10^{-3}$ & 13.70 & \cellcolor{blue!10}$\mathbf{2.10 \times 10^{-5}}$ & \cellcolor{blue!10}$\mathbf{3.12 \times 10^{-8}}$ & 5.30 & $4.48 \times 10^{0}$ & $3.70 \times 10^{-3}$ & 17.40 & $1.92 \times 10^{1}$ & $6.67 \times 10^{-2}$ \\
10   & \cellcolor{blue!10}\textbf{2.41} & \cellcolor{blue!10}$\mathbf{1.40 \times 10^{-6}}$ & \cellcolor{blue!10}$\mathbf{9.56 \times 10^{-9}}$ & 13.48 & $2.10 \times 10^{-5}$ & $3.12 \times 10^{-8}$ & 5.52 & $6.74 \times 10^{-4}$ & $1.08 \times 10^{-6}$ & 16.36 & $9.60 \times 10^{0}$ & $3.66 \times 10^{-2}$ \\
100  & \cellcolor{blue!10}\textbf{2.52} & \cellcolor{blue!10}$\mathbf{1.67 \times 10^{-6}}$ & \cellcolor{blue!10}$\mathbf{8.97 \times 10^{-9}}$ & 13.16 & $2.10 \times 10^{-5}$ & $3.12 \times 10^{-8}$ & 7.57 & $1.61 \times 10^{-4}$ & $8.87 \times 10^{-7}$ & 19.75 & $1.65 \times 10^{1}$ & $7.24 \times 10^{-2}$ \\
1000 & \cellcolor{blue!10}\textbf{2.59} & \cellcolor{blue!10}$\mathbf{1.67 \times 10^{-6}}$ & \cellcolor{blue!10}$\mathbf{6.10 \times 10^{-9}}$ & 12.92 & $2.10 \times 10^{-5}$ & $3.12 \times 10^{-8}$ & 11.32 & $5.17 \times 10^{-5}$ & $6.43 \times 10^{-7}$ & 48.68 & $2.05 \times 10^{1}$ & $6.03 \times 10^{-2}$ \\
2000 & \cellcolor{blue!10}\textbf{2.73} & \cellcolor{blue!10}$\mathbf{1.66 \times 10^{-6}}$ & \cellcolor{blue!10}$\mathbf{9.70 \times 10^{-9}}$ & 13.37 & $2.10 \times 10^{-5}$ & $3.12 \times 10^{-8}$ & 14.62 & $3.96 \times 10^{-5}$ & $5.94 \times 10^{-7}$ & 87.89 & $1.52 \times 10^{1}$ & $4.87 \times 10^{-2}$ \\
\hline
\end{tabular}%
}
\vspace{-5pt}
\end{table*}

%% file: tex/figs/fig-colfreeTable.tex
\begin{table*}[t]
\centering
\large
\setlength{\tabcolsep}{4pt}
\caption{Batch-size comparison across solvers for the Panda Arm, evaluated on collision-free solutions in the MotionBenchMaker \textit{box\_panda} scenario. Values are the averages across all solves. Position error is in millimeters, orientation error is in radians, and success rate denotes the percentage of queries for which a collision-free IK solution was found. HJCD-IK demonstrates as much as order-of-magnitude improvements.}
\label{tab:ik_colfree_box}
\vspace{-4pt}
\resizebox{\textwidth}{!}{%
\begin{tabular}{r|cccc|cccc|cccc}
\hline
Batch & \multicolumn{4}{c|}{HJCD-IK} & \multicolumn{4}{c|}{PyRoki} & \multicolumn{4}{c|}{cuRobo} \\
\cline{2-10}
\hline
& Time (ms) & Pos. Err. (mm) & Ori. Err. (rad) & Succ.(\%)
& Time (ms) & Pos. Err. (mm) & Ori. Err. (rad) & Succ.(\%)
& Time (ms) & Pos. Err. (mm) & Ori. Err. (rad) & Succ.(\%) \\
\hline
1    & \cellcolor{blue!10}\textbf{5.44} & $8.17$ & $1.96 \times 10^{-2}$ & 89.0 & 34.04 & $5.18 \times 10^{2}$ & $3.96 \times 10^{-1}$ & 6.0 & 23.76 & \cellcolor{blue!10}$\mathbf{7.85}$ & \cellcolor{blue!10}$\mathbf{2.61 \times 10^{-3}}$ & \cellcolor{blue!10}\textbf{97.0} \\
10   & \cellcolor{blue!10}\textbf{4.19} & $7.11 \times 10^{-4}$ & $5.34 \times 10^{-7}$ & 98.0 & 46.29 & \cellcolor{blue!10}$\mathbf{9.90 \times 10^{-5}}$ & \cellcolor{blue!10}$\mathbf{1.69 \times 10^{-7}}$ & 89.0 & 29.31 & $2.43 \times 10^{-3}$ & $4.00 \times 10^{-6}$ & \cellcolor{blue!10}\textbf{100.0} \\
100  & \cellcolor{blue!10}\textbf{4.42} & \cellcolor{blue!10}$\mathbf{8.83 \times 10^{-5}}$ & \cellcolor{blue!10}$\mathbf{3.56 \times 10^{-8}}$ & \cellcolor{blue!10}\textbf{100.0} & 48.98 & $9.80 \times 10^{-5}$ & $1.41 \times 10^{-7}$ & 93.0 & 30.76 & $6.99 \times 10^{-4}$ & $2.00 \times 10^{-6}$ & \cellcolor{blue!10}\textbf{100.0} \\
1000 & \cellcolor{blue!10}\textbf{5.04} & \cellcolor{blue!10}$\mathbf{2.03 \times 10^{-5}}$ & \cellcolor{blue!10}$\mathbf{9.06 \times 10^{-9}}$ & \cellcolor{blue!10}\textbf{100.0} & 46.98 & $8.70 \times 10^{-5}$ & $1.36 \times 10^{-7}$ & 92.0 & 28.50 & $2.93 \times 10^{-4}$ & $2.00 \times 10^{-6}$ & \cellcolor{blue!10}\textbf{100.0} \\
2000 & \cellcolor{blue!10}\textbf{5.35} & \cellcolor{blue!10}$\mathbf{1.71 \times 10^{-5}}$ & \cellcolor{blue!10}$\mathbf{7.16 \times 10^{-9}}$ & \cellcolor{blue!10}\textbf{100.0} & 35.74 & $8.80 \times 10^{-5}$ & $1.49 \times 10^{-7}$ & 90.0 & 61.96 & $2.47 \times 10^{-4}$ & $2.00 \times 10^{-6}$ & \cellcolor{blue!10}\textbf{100.0}\\
\hline
\end{tabular}%
}
\vspace{-5pt}
\end{table*}

%% file: tex/figs/fig-dofTable.tex
\begin{table*}[t]
\centering
\small
\setlength{\tabcolsep}{4pt}
\caption{DoF scalability for HJCD-IK, PyRoki, and cuRobo with batch size of 1000, demonstrating HJCD-IK's improved performance.}
\label{tab:ik_dof}
\vspace{-4pt}
\resizebox{\textwidth}{!}{%
\begin{tabular}{r|ccc|ccc|ccc}
\hline
DoF & \multicolumn{3}{c|}{HJCD-IK} & \multicolumn{3}{c|}{PyRoki} & \multicolumn{3}{c}{cuRobo} \\
\cline{2-10}
& Time (ms)  & Pos. Err. (mm)   & Ori. Err. (rad)  
& Time (ms)   & Pos. Err. (mm)   & Ori. Err. (rad)  
& Time (ms)   & Pos. Err. (mm)   & Ori. Err. (rad)   \\
\hline
7  & \cellcolor{blue!10}\textbf{4.25} & \cellcolor{blue!10}$\mathbf{1.71 \times 10^{-5}}$ & \cellcolor{blue!10}$\mathbf{4.11 \times 10^{-8}}$ & 15.09 & $2.63 \times 10^{-2}$ & $3.70 \times 10^{-5}$ & 9.11 & $3.38 \times 10^{-4}$ & $1.59 \times 10^{-6}$ \\
12 & \cellcolor{blue!10}\textbf{4.55} & \cellcolor{blue!10}$\mathbf{1.94 \times 10^{-5}}$ & \cellcolor{blue!10}$\mathbf{6.91 \times 10^{-8}}$ & 16.29 & $1.99 \times 10^{-2}$ & $1.86 \times 10^{-5}$ & 12.66 & $7.78 \times 10^{-1}$ & $2.57 \times 10^{-2}$ \\
18 & \cellcolor{blue!10}\textbf{4.62} & \cellcolor{blue!10}$\mathbf{3.76 \times 10^{-5}}$ & \cellcolor{blue!10}$\mathbf{6.95 \times 10^{-8}}$ & 20.82 & $2.15 \times 10^{-2}$ & $2.14 \times 10^{-5}$ & 16.26 & $8.41 \times 10^{-1}$ & $3.03 \times 10^{-2}$ \\
24 & \cellcolor{blue!10}\textbf{4.66} & \cellcolor{blue!10}$\mathbf{3.84 \times 10^{-5}}$ & \cellcolor{blue!10}$\mathbf{7.32 \times 10^{-8}}$ & 24.34 & $1.84 \times 10^{-2}$ & $1.99 \times 10^{-5}$ & 19.55 & $7.50 \times 10^{-1}$ & $3.58 \times 10^{-2}$ \\
\hline
\end{tabular}%
}
\vspace{-15pt}
\end{table*}

%% file: tex/figs/fig-colfreeStudy.tex
\iffigs
\begin{figure}[!ht]
    \centering
    \includegraphics[width=\columnwidth]{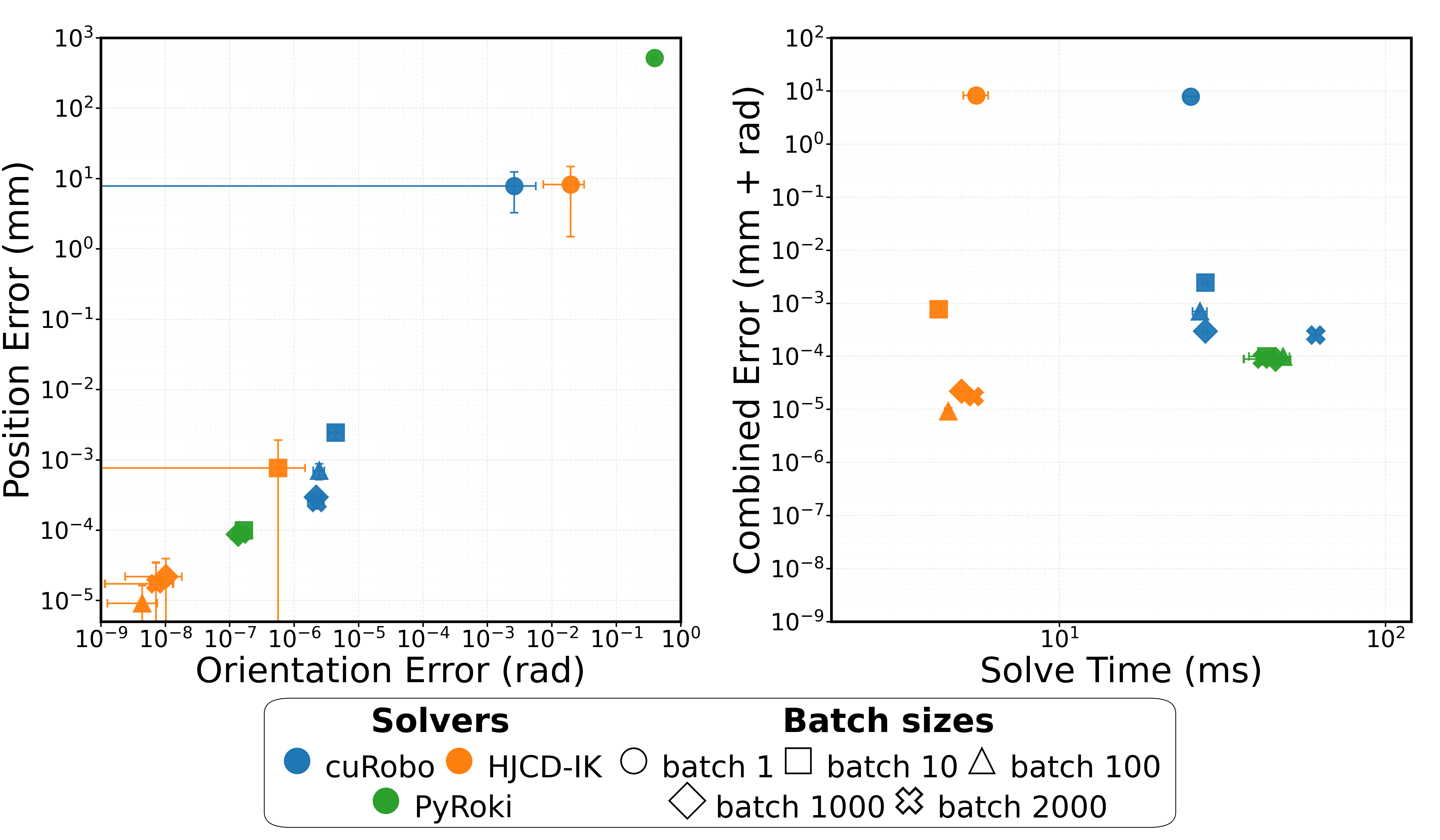}
    \vspace{-15pt}
    \caption{Position vs Orientation Error and Combined Error (mm and radians) of HJCD-IK (orange), cuRobo (blue), and PyRoki (green) on the Panda evaluated on collision-free solutions in the MotionBenchMaker \textit{box\_panda} scenario (data in~\cref{tab:ik_colfree_box}). Across batch sizes, HJCD-IK achieves the lowest latency while remaining on or near the accuracy-latency Pareto frontier.}
    \label{fig:meta_colfree}
    \vspace{-15pt}
\end{figure}
\fi

%% file: tex/figs/fig-dofScalingStudy.tex
\iffigs
\begin{figure}[!ht]
    \centering
    \includegraphics[width=\columnwidth]{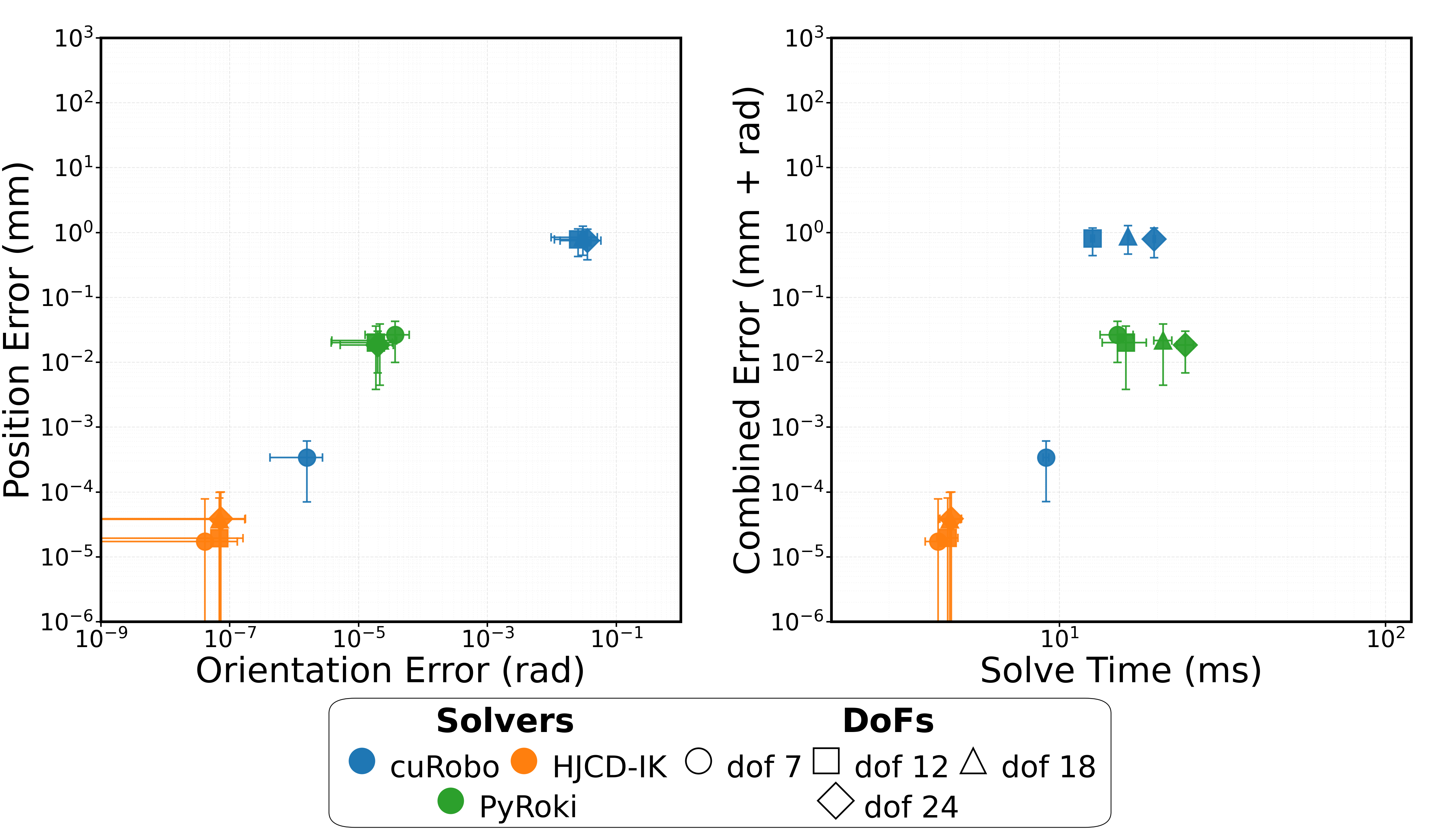}
    \vspace{-15pt}
    \caption{Position vs. Orientation Error and Combined Error (mm and radians) of HJCD-IK (orange), cuRobo (blue), and PyRoki (green) while varying the DoF of the synthetic arm under analysis in an open-world random target benchmark (data in~\cref{tab:ik_dof}). Across all DoF, HJCD-IK outperforms all baselines on both error and latency.}
    \label{fig:meta_dof}
    \vspace{-10pt}
\end{figure}
\fi

%% file: tex/figs/fig-hardware.tex
\iffigs
\begin{figure}[!ht]
    \centering
    \includegraphics[width=0.75\columnwidth]{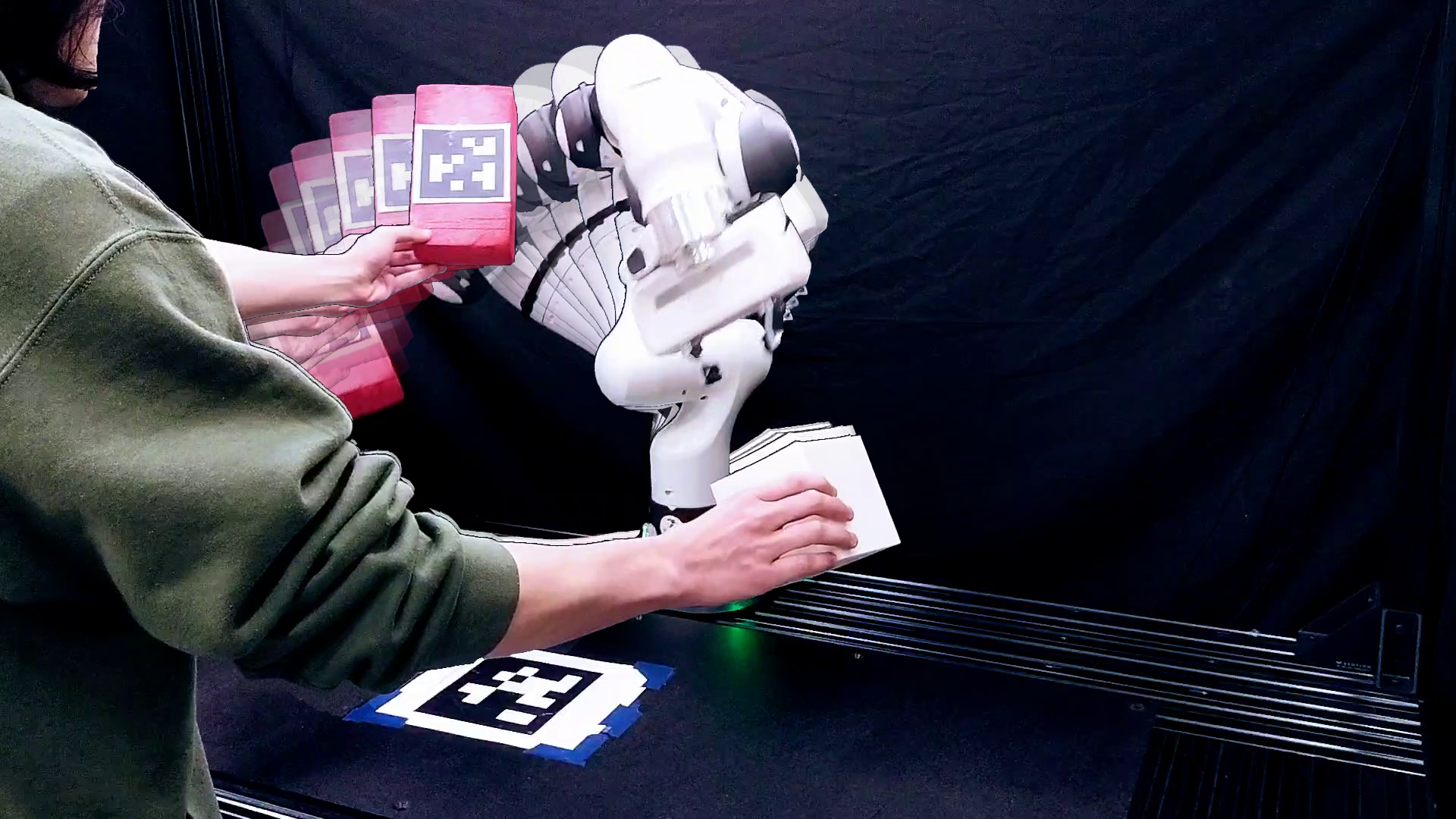}
    \caption{Dynamic Tracking and Collision Avoidance using HJCD-IK}
    \label{fig:hardware}
\end{figure}
\fi

%% file: tex/figs/fig-mmdTable.tex
\begin{table}[!t]
\centering
\caption{MMD and MMD$^2$ across solvers for the Panda Arm.}
\label{tab:mmd}
\vspace{-4pt}
\begin{tabular}{lrrrr}
\toprule
Metric & \textbf{HJCD-IK} & PyRoki & cuRobo & IKFlow \\
\midrule
MMD $\downarrow$ & \cellcolor{blue!10}\textbf{0.02261} & 0.04514 & 0.05348 & 0.03670 \\
MMD$^2$ $\downarrow$ & \cellcolor{blue!10}\textbf{0.00051} & 0.00203 & 0.00286 & 0.00134 \\
\bottomrule
\end{tabular}
\vspace{-10pt}
\end{table}